%% file: main.tex
\documentclass{article}

\PassOptionsToPackage{numbers, compress}{natbib}
 \usepackage[main, final]{neurips_2025}

\usepackage[utf8]{inputenc} 
\usepackage[T1]{fontenc}    
\usepackage{hyperref}       
\usepackage{url}            
\usepackage{booktabs}       
\usepackage{amsfonts}       
\usepackage{nicefrac}       
\usepackage{microtype}      
\input{preamble}

\title{Unmasking Puppeteers: Leveraging Biometric Leakage to Expose Impersonation in AI-based Videoconferencing}

\author{
\textbf{Danial Samadi Vahdati\thanks{Corresponding author: ds3729@drexel.edu}\textsuperscript{\rm 1}}, \
\textbf{Tai Duc Nguyen\textsuperscript{\rm 1}}, \
\textbf{Koki Nagano\textsuperscript{\rm 2}}, \\
\textbf{David Luebke\textsuperscript{\rm 2}}, \
\textbf{Orazio Gallo\textsuperscript{\rm 2}}, \
\textbf{Ekta Prashnani\textsuperscript{\rm 2}} \\
\textbf{Matthew Stamm\textsuperscript{\rm 1}} \\
\textsuperscript{\rm 1}Drexel University, \
\textsuperscript{\rm 2}NVIDIA
}

\begin{document}

\maketitle

\input{sec/Abstract_Tai_v1}

\input{sec/intro_dsv_trimm}
\input{sec/Background_dsv_trim}

\input{sec/Problem_Formulation_v5_MCS}

\input{sec/proposed_approach_v9_dsv_rebutt}



\input{sec/experiments_dsv_trim_v4}

\input{sec/ablation_dsv_v6_trim_rebut}
\input{sec/discussion_dsv_trim}

\input{sec/Conclusion_Tai_v1}

\section*{Acknowledgments}
This material is based upon work supported by the National Science Foundation under Grant No. 2320600. Any opinions, findings, and conclusions or recommendations expressed in this material are those of the authors and do not necessarily reflect the views of the National Science Foundation.This research was supported in part by NVIDIA Corporation. We thank the NVIDIA Research team for their collaboration, resources, and valuable feedback throughout this project.

{
\small
\bibliographystyle{unsrtnat}
\bibliography{main}
}

\newpage
\include{check_list}

\newpage
\include{supp_mat}

\end{document}

%% file: preamble.tex
\usepackage[dvipsnames]{xcolor}
\usepackage{amsmath}
\usepackage{verbatim}
\usepackage{tabularray}
\usepackage{array}
\usepackage{rotating}
\usepackage{multirow}
\usepackage{ragged2e}
\usepackage{scalerel}
\usepackage{bbm}
\usepackage{graphicx}
\usepackage{wrapfig}
\usepackage{booktabs}
\usepackage{xspace}
\usepackage{fontawesome}
\PassOptionsToPackage{numbers, compress}{natbib}
\usepackage{neurips_2025}
\usepackage{subcaption}
\usepackage{float}
\usepackage{enumitem}

\newcommand{\red}[1]{{\color{red}#1}}
\newcommand{\blue}[1]{{\color{blue}#1}}
\newcommand{\subhead}[1]{\vspace{0.25\baselineskip} \noindent \textbf{#1.}}
\newcommand{\subheader}[1]{\vspace{0.25\baselineskip} \noindent \textbf{#1.}}

\newcommand{\pullup}{\vspace{-0.2\baselineskip}}
\newcommand{\pullupp}{\vspace{-0.4\baselineskip}}
\newcommand{\pulluppp}{\vspace{-0.8\baselineskip}}

%% file: sec/Abstract_Tai_v1.tex
\begin{abstract}
    AI-based talking-head videoconferencing systems reduce bandwidth by sending a compact pose-expression latent and re-synthesizing RGB at the receiver—but this latent can be “puppeteered,” letting an attacker hijack a victim’s likeness in real time. Because every frame is synthetic, deepfake and synthetic video detectors fail outright.
    To address this security problem, we exploit a key observation: the pose expression latent inherently contain biometric information of the driving identity. Therefore, we introduce the first \emph{biometric leakage defense} without ever looking at the reconstructed RGB video: a pose-conditioned, large-margin contrastive encoder that isolates persistent identity cues inside the transmitted latent while cancelling transient pose and expression. A simple cosine test on this disentangled embedding flags illicit identity swaps as the video is rendered. Our experiments on multiple talking-head generation models show that our method consistently outperforms existing puppeteering defenses, operates in real-time, and shows strong generalization to out-of-distribution scenarios.Our code and trained models are available at \href{https://github.com/MISLresearch/Unmasking-Puppeteers-Neurips25}{https://github.com/MISLresearch/Unmasking-Puppeteers-Neurips25}.

\end{abstract}

%% file: sec/intro_dsv_trimm.tex
\section{Introduction}

Advances in generative AI have enabled the creation of hyper-realistic synthetic videos. 
This has led to the development of many new technologies, including both avatar-based communication systems~\cite{chu2020expressive,li2025generating,van2022deep} and AI-based videoconferencing systems~\cite{stengel2023ai, rajaram2024blendscape, li2022faivconf, cutler2020multimodal}. 
AI-based talking head videoconferencing systems are receiving increasing attention due to their significant bandwidth savings~\cite{wang2021one}. Instead of continuously encoding and transmitting each video frame of a speaker, these systems only transmit embeddings that capture a speaker's pose and facial expression.  Then, a generative AI system at the receiver's end uses these embeddings, in conjunction with an initial representation of the speaker, to create an accurately reconstructed video.

Unfortunately, AI-based talking head videoconferencing systems~\cite{rajaram2024blendscape,stengel2023ai,wang2021one,cutler2020multimodal,tong2024multimodal} are vulnerable to a new form of information attack known as ``puppeteering''~\cite{vahdati2023defending}. In this attack, a malicious user at the sender side transmits an unauthorized representation of a different target speaker when a video call is initiated. As a result, the identity in the video that the receiver constructs is different than the identity of the person who actually controls the video~\cite{vahdati2023defending,prashnani2024avatar}.

\begin{figure}[!t]
\centering
\begin{minipage}{0.85\linewidth}
    \centering
    \includegraphics[width=0.85\linewidth]{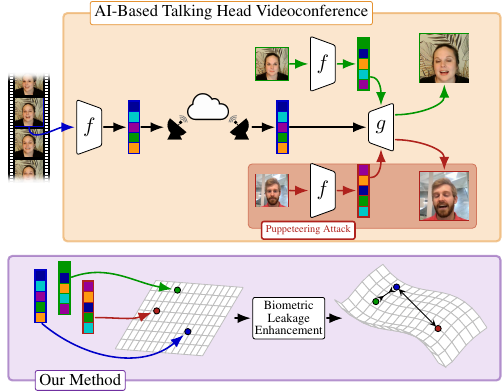}
\end{minipage}%
\hfill
\begin{minipage}{\linewidth}
    \captionof{figure}{\label{fig:teaser}
    AI-based talking-head generators transmit only a compact pose-and-expression embedding for low-bandwidth videoconferencing, but remain vulnerable to puppeteering attacks that swap in a different identity for live impersonation. Our defense capitalizes on biometric signals inadvertently leaked in these embeddings to reveal mismatches between the driving speaker and the reconstructed identity in real time.}
\end{minipage}
\vspace{-10pt}
\end{figure}

While many forensic approaches, such as deepfake~\cite{coccomini2022combining,bonettini2021video,xu2023tall,xu2022supervised,nguyen2024laa,nguyen2024videofact} and synthetic video detectors~\cite{vahdati2024beyond,ji2024distinguish,chang2024what,li2024towards}, aim to expose unauthorized AI-generated media, they operate by identifying evidence that a video has been synthetically generated. 
However, in AI-based talking-head videoconferencing, \emph{every video is AI-generated}, rendering existing media forensic approaches unable to detect puppeteering.

Puppeteering attacks pose a fundamentally different threat from those typically addressed by media forensics. Instead of asking ``Is this video real or AI-generated?'' one key question must be 
``Is this AI-generated video being driven by an authorized identity, or has someone hijacked it?''
As talking-head videoconferencing technology continues to advance, it is essential to mitigate puppeteering attacks before these systems are widely deployed. Addressing these threats proactively will help ensure secure and trustworthy communication over these systems.

Few defenses address puppeteering. Prashnani et al. introduced Avatar Fingerprinting~\cite{prashnani2024avatar}, which works by enrolling each user’s characteristic facial motion patterns into the detector's training data. Although this could help, it requires significant user-specific data to build reliable motion signatures. In contrast, Vahdati et al. avoid enrollment by decoding and re-encoding the received video for comparison \cite{vahdati2023defending}; the round-trip is compute-heavy and adds distortions that hurt accuracy.

In this paper we introduce a real-time, enrollment-free defense that detects puppeteering attacks entirely in the latent domain. Our key insight is that the pose-and-expression embeddings already transmitted by modern talking-head systems leak subtle, but reproducible, biometric signatures~\cite{chu2024generalizable, danvevcek2022emoca}. We learn a compact Enhanced Biometric-Leakage (EBL) space in which identity cues are amplified while pose and expression variance is actively suppressed. A pose-conditioned contrastive loss drives this separation, and a lightweight temporal LSTM aggregates evidence to yield stable, millisecond-level decisions. By comparing the sender’s latent identity with the target identity rendered at the receiver, our system flags any mismatch without decoding RGB frames, tracking landmarks, or collecting per-user motion profiles. Extensive experiments across many generators and datasets show that our method achieves state-of-the-art detection performance while working in real-time.

\noindent In summary, our contributions are as follows.
\vspace{-4pt}
\begin{itemize}[leftmargin=0.6cm]
    \item We present the first method that operates solely on data already available at the receiver, requiring no user enrollment or additional sensors.
    \item We formalize and harness identity leakage inherent in pose–expression embeddings, turning a previously ignored vulnerability into a defensive signal.
    \item We develop a novel loss and training protocol to learn a low-dimensional EBL representation that maximizes identity separability while nullifying pose and expression.
    \item Though extensive experiments across fifteen generator/dataset combinations, our approach achieves state-of-the-art puppeteering detection in real time, demonstrating practical viability for deployment in bandwidth-constrained videoconferencing systems.
\end{itemize}
\vspace{-8pt}

%% file: sec/Background_dsv_trim.tex
\section{Background and Related Work}
\label{sec:background}
\subheader{Talking-Head Video Systems}
Talking-head systems synthesize facial motion and speech in real time by transmitting low-dimensional embeddings of pose and expression rather than full video frames~\cite{wang2021audio2head,shen2024emotalker,oorloff2022expressive,chen2020talking}. A generator at the receiver reconstructs a realistic face. Methods range from compressed embeddings to landmark-based updates per frame~\cite{wang2021one,agarwal2021compressing,hong2022depth,hong2023dagan++,mallya2022implicit,wu2022anifacegan,song2022talking}, reducing bandwidth while maintaining high visual fidelity~\cite{gao2023high,hwang2023discohead,wang2024emotivetalk,wang2023interactive,ye2023r2talker,oquab2022efficient,tandon2022txt2vid}. Systems like NVIDIA’s Vid2Vid Cameo~\cite{wang2021one} and Google’s Project Starline~\cite{lawrence2021project} demonstrate real-time, 3D teleconferencing.

\subheader{Forensics for Synthetic Media}
Synthetic media forensics traditionally aims to distinguish real from AI-generated content. Deepfake detectors~\cite{coccomini2022combining,bonettini2021video,xu2023tall,xu2022supervised,nguyen2024laa,nguyen2024videofact,knafo2022fakeout} leverage semantic cues~\cite{hu2021exposing,li2018detecting,li2020deeprhythm,demir2021fakecatcher}, lighting inconsistencies~\cite{wu2022capturing,tian2024illumination,lai2024lideepdet,ciamarra2023surfake}, and facial texture artifacts~\cite{chen2023aim,guo2023exposing,li2022artifacts}, while synthetic video detectors~\cite{vahdati2024beyond,ji2024distinguish,chang2024what,li2024towards} focus on identifying AI-generated traces. However, in low-bandwidth talking-head systems, every video is synthetic~\cite{vahdati2023defending,prashnani2024avatar}, rendering these detectors ineffective - they cannot determine whether the video represents the correct identity or has been hijacked.

\subheader{Puppeteering Attacks and Existing Defenses}
Puppeteering attacks replace the target’s appearance with the adversary’s pose-and-expression vectors at call initiation, causing the receiver to synthesize a video of the wrong identity.  
Avatar Fingerprinting~\cite{prashnani2024avatar} mitigates this threat by first enrolling each user’s facial-motion signatures; at inference it flags a session whenever the incoming motion deviates from the stored profile.  
Vahdati et al.~\cite{vahdati2023defending} avoid enrollment by re-encoding the synthesized video and comparing embeddings, but their pipeline depends on facial-landmark estimation, which degrades under large head rotations and lighting artifacts introduced by video reconstruction.

In contrast, our method exploits biometric leakage \emph{already present} in the pose-and-expression latents, removing both the enrollment phase and any reliance on facial landmarks.
A contrastive objective disentangles identity from pose, yielding a compact, latency-friendly representation that generalizes to unseen generators and real-world conferencing conditions.

\begin{figure*}[!t]
    \centering
    \resizebox{\linewidth}{!}{%
    \begin{tabular}{ccc|ccc|ccc}
        \multicolumn{3}{c|}{\huge\textbf{NVIDIA-VC}} & 
        \multicolumn{3}{c|}{\huge\textbf{RAVDESS}} & 
        \multicolumn{3}{c}{\huge\textbf{CREMA-D}} \\[6pt]
        
        {\Large\textbf{Reference}} & {\Large\textbf{Self-Reenacted}} & {\Large\textbf{Cross-Reenacted}} & 
        {\Large\textbf{Reference}} & {\Large\textbf{Self-Reenacted}} & {\Large\textbf{Cross-Reenacted}} & 
        {\Large\textbf{Reference}} & {\Large\textbf{Self-Reenacted}} & {\Large\textbf{Cross-Reenacted}} \\[6pt]
        
        \includegraphics[width=0.190\textwidth]{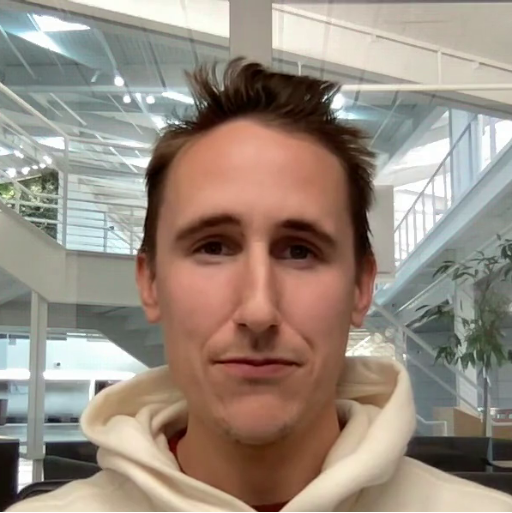} & 
        \includegraphics[width=0.190\textwidth]{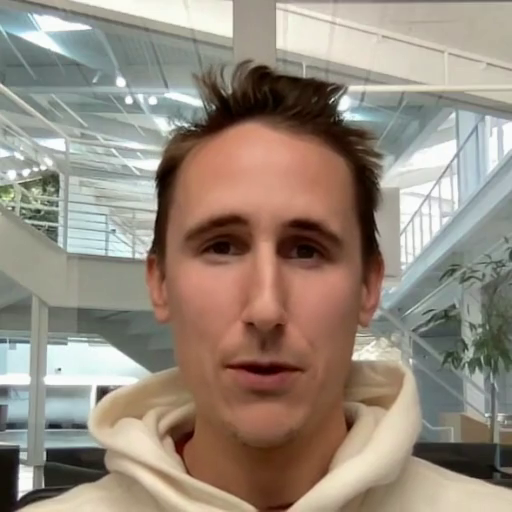} & 
        \includegraphics[width=0.190\textwidth]{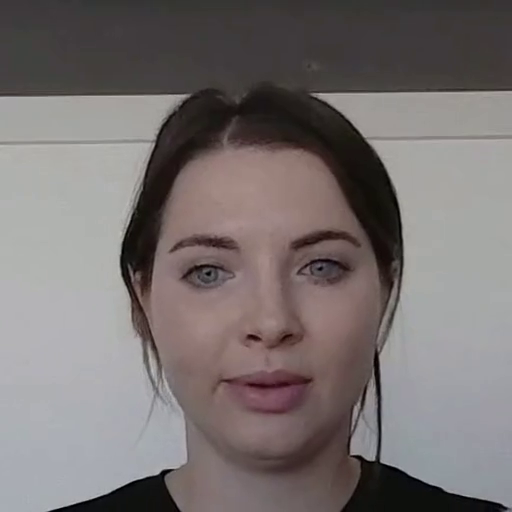} &
        \includegraphics[width=0.190\textwidth]{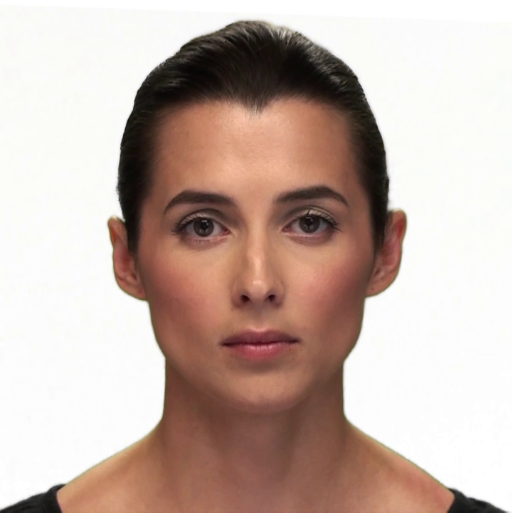} & 
        \includegraphics[width=0.190\textwidth]{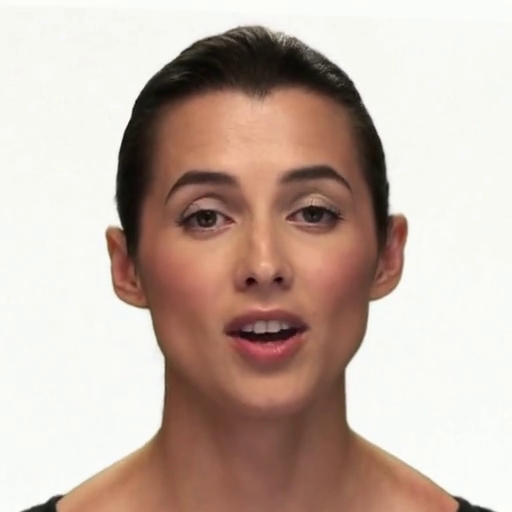} & 
        \includegraphics[width=0.190\textwidth]{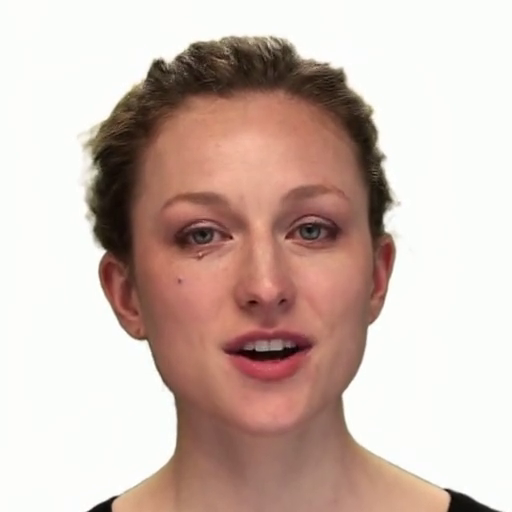} &
        \includegraphics[width=0.190\textwidth]{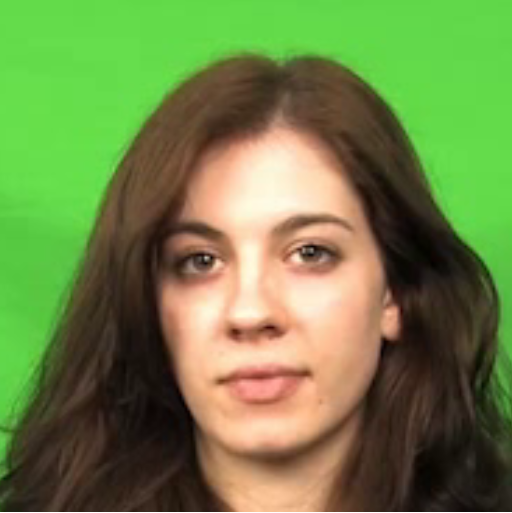} & 
        \includegraphics[width=0.190\textwidth]{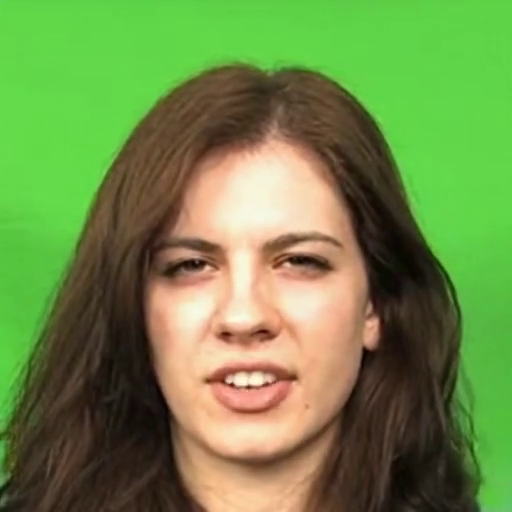} & 
        \includegraphics[width=0.190\textwidth]{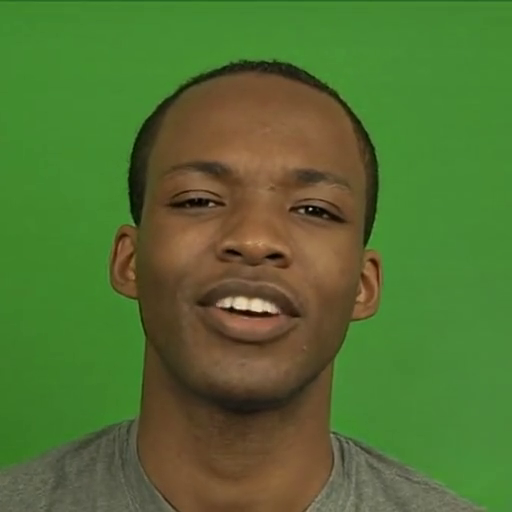} \\
    \end{tabular}
    }
    \caption{Illustration of three datasets (NVIDIA-VC~\cite{prashnani2024avatar}, RAVDESS~\cite{livingstone2018ryerson}, CREMA-D~\cite{cao2014crema}) shown across three columns each. Row 2 indicates the type of frame displayed in Row 3: Reference, Self-Reenacted, or Cross-Reenacted.}
    \label{fig:three_datasets}
    \pulluppp\pullupp
\end{figure*}

%% file: sec/Problem_Formulation_v5_MCS.tex

\section{Problem Formulation}
\label{sec:methodology}


\subsection{Talking Head Video Systems}

%
AI-enabled talking-head videoconferencing systems 
have been proposed 
to  minimize transmitted data by encoding and transmitting only pose and expression information from a speaker rather than an entire video frame~\cite{wang2021one, zhao2024synergizing, tan2024say, ma2024cvthead, shen2022learning, zhang2023sadtalker, wang2022latent, agarwal2023audio}. As this technology rapidly matures, our goal here is to develop security measures that improve the trustworthiness of these systems when they are deployed.

A videoconferencing call starts with each participant $k$ sending a neutral reference portrait $R^k$ to the receiver.  
Typically, $R^k$ is an image of the speaker in a neutral pose. As the speaker continues to speak, each new video frame $V^k_t$ capturing this speaker at the sender side is processed by an embedding function $f$ to produce an embedding $z^k_t=f(V^k_t)$ that encodes the speaker's instantaneous pose and expression at time $t$.  This embedding is transmitted to the receiver, as is shown in Fig.~\ref{fig:teaser}.

At the receiver side, a generator $g$ is used to produce a video of the speaker on the basis of the received embedding $z_t^k$ and the reference representation $R^k$ such that
\begin{equation}
\hat{V}_t^{k\rightarrow k} = g(z_t^k,R^k).
\end{equation}
Here, the superscript $\hat{V}_t^{k\rightarrow k}$ denotes that speaker $k$ both \emph{drives and appears} in the rendered video.
We refer to the sender’s identity as the ``driving identity'' and the reference portrait sent to the receiver as the ``target identity''~\cite{prashnani2024avatar}.
While such a talking-head video-conferencing system promises to be bandwidth-efficient, it exposes a vulnerability to puppeteering attacks when a malicious sender impersonates another individual.

\subsection{Puppeteering Attacks}

A puppeteering attack exploits the inherent trust in the reference representation transmitted to the receiver when a videoconferencing call begins. Here, an adversary, speaker $\ell$, obtains a target speaker’s representation $R^{k}$ and substitutes it for their own without authorization at the start of the call~\cite{vahdati2023defending}. 
As the call proceeds, the adversary’s own video $V^{\ell}_t$ is used to derive the pose and expression vectors $z^{\ell}_t$ that are transmitted to the receiver. The receiver then uses the generator $g$ alongside the unauthorized reference $R^{k}$ and $z^{\ell}_t$ to reconstruct the video.  As a result, the receiver is presented with a realistic-looking video of speaker $k$, controlled in real time by the speaker $\ell$ such that 
\begin{equation}
    \hat{V}_t^{k \rightarrow \ell} = g\bigl(f(V^{\ell}_t), R^{k}\bigr) = g\bigl(z^{\ell}_t, R^{k}\bigr),
\end{equation}
where $\hat{V}_t^{k \rightarrow \ell}$ is an unauthorized impersonation of speaker $\ell$ driven by speaker $k$. 
To address this threat, one must come up with mechanisms to detect whether the receiver's generated video is driven by the same authorized identity, rather than an unauthorized impersonator.


\subsection{Why Real-vs-Synthetic Detectors Fail}

Deepfake and synthetic-video detectors flag frames whose pixel statistics reveal AI generation, implicitly assuming that “real” camera footage is the baseline. In bandwidth-efficient talking-head conferencing, however, every frame - even those from honest participants - is produced by a generator; “synthetic” is the norm, not the anomaly. The security question therefore shifts from ``Was this video AI-generated?'' to ``Does the rendered face correspond to the person actually driving the latents?'' Pixel-level detectors cannot answer that mapping, leaving puppeteering attacks undetectable.

%% file: sec/proposed_approach_v9_dsv_rebutt.tex
\section{Proposed Approach}

In this paper, we present a real-time solution for detecting puppeteering attacks in talking-head videoconferencing, uniquely operating entirely in the latent domain without access to reconstructed RGB frames. To do this, we re-encode each pose-and-expression latent embedding into a compact Enhanced Biometric Leakage (EBL) embedding space that captures who is speaking while discarding how they move. A pose-conditioned contrastive loss learns this space by pulling together identity-consistent pairs and pushing apart identity-mismatched, pose-matched pairs. At run time, we simply compare each live EBL vector to the EBL embedding of the reference portrait sent at call setup; a sharp drop in similarity reveals that the driving and target identities have diverged, signalling puppeteering. For additional robustness and stable authentication, during training, we discard unreliable extreme-pose frames, and leverage an LSTM to fuse successive EBL scores to output stable, low-latency decisions. These innovations yield an algorithm that authenticates speakers in real-time, requires no enrollment or facial-landmark preprocessing. The following sections detail our approach and its implementation.

\begin{figure}[!t]
    \centering
    \begin{minipage}[b]{0.48\linewidth}
        \centering
        \includegraphics[height=3.2cm]{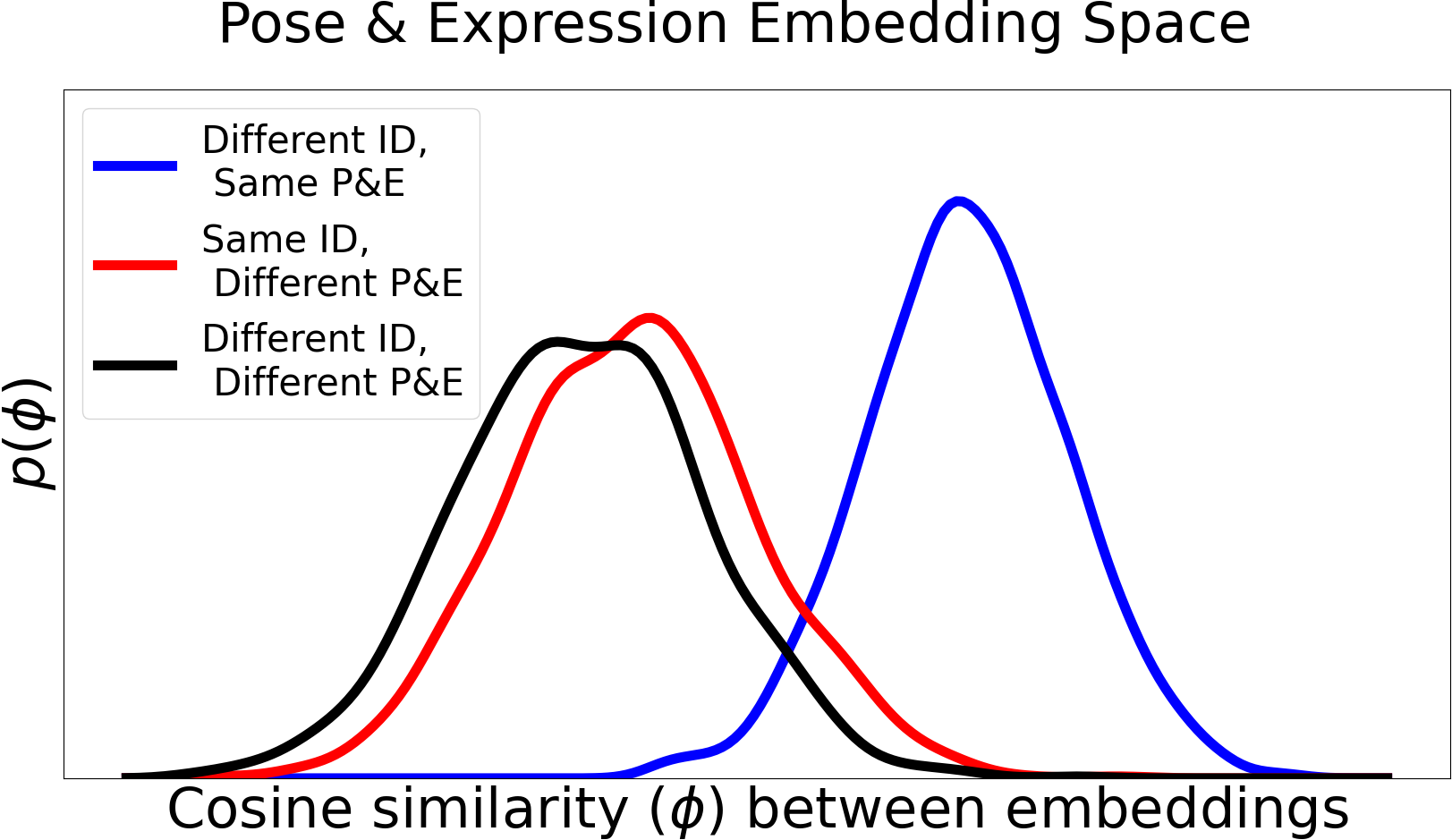}  
    \end{minipage}
    \hfill
    \begin{minipage}[b]{0.48\linewidth}
        \centering
        \includegraphics[height=3.2cm]{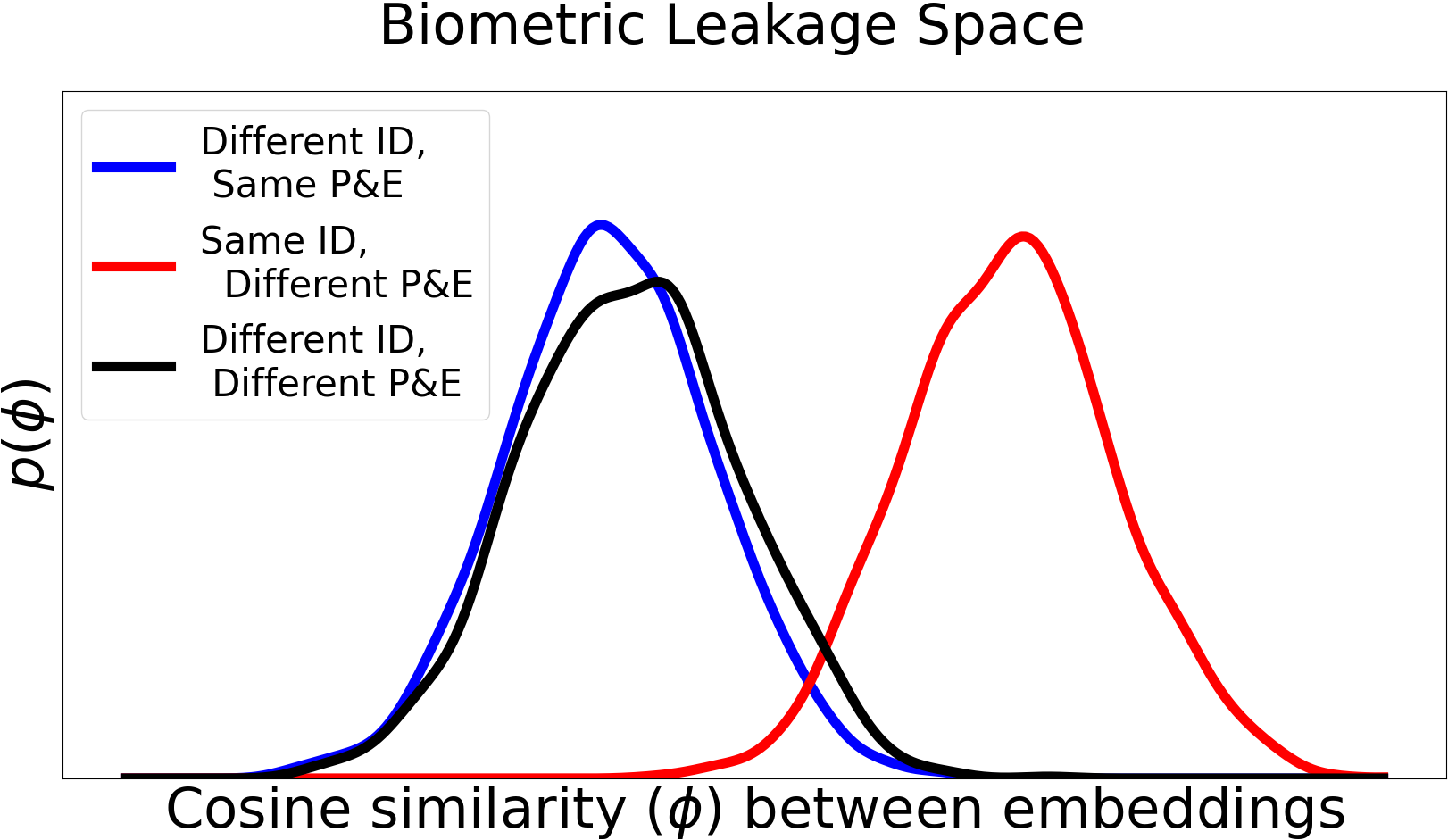}  
    \end{minipage}
    \vspace{1.5mm}
    \begin{minipage}[b]{0.32\linewidth}
        \centering
        \includegraphics[height=2.3cm]{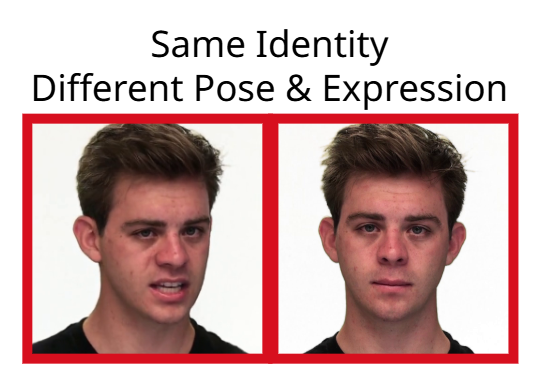}  
    \end{minipage}
    \hfill
    \begin{minipage}[b]{0.32\linewidth}
        \centering
        \includegraphics[height=2.3cm]{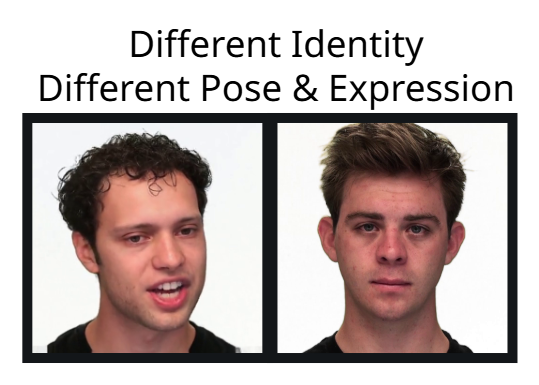}  
    \end{minipage}
    \hfill
    \begin{minipage}[b]{0.32\linewidth}
        \centering
        \includegraphics[height=2.3cm]{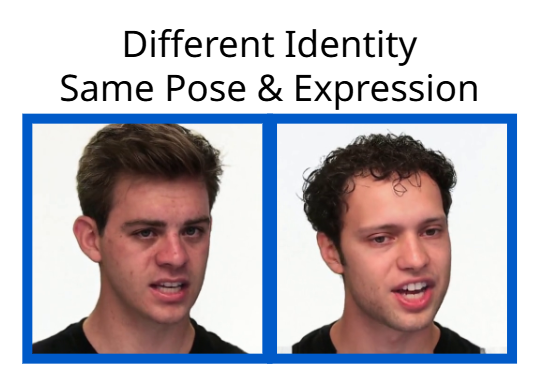}  
    \end{minipage}
    \caption{\label{fig:diff_distrib}Similarity distributions in P\&E space (left) and biometric leakage space (right). \red{Red}: same ID, diff. P\&E; \blue{blue}: diff. ID, same P\&E; black: diff. ID, diff. P\&E.}
    \pulluppp
\end{figure}

\subsection{Biometric Entanglements in Pose \& Expression Latent Space}

Low-bandwidth talking-head systems encode each frame $V_t$ as a pose-and-expression vector $z_t = f(V_t)$, which the receiver’s generator turns back into pixels. Although designed for \emph{geometry}, $z_t$ inherit identity cues from the physical face that produced the motion. Measurements such as inter-ocular distance, jaw curvature, or lip thickness inevitably contaminate $z_t$ because head pose and facial musculature cannot be sensed independently of the speaker’s anatomy. Prior studies to disentangle identity from pose and expression also reported persistent identity traces in pose-conditioned representations~\cite{chu2024generalizable, danvevcek2022emoca}, which provides the empirical basis for our proposed defense.

Let $R$ denote the static reference portrait of the nominal speaker transmitted at session start. Both endpoints apply the shared encoder $f$ to compute the reference embedding $f(R)$, placing $R$ and the live stream $\{z_t\}$ in one latent space. A natural baseline defense is to compute the cosine similarity $s_c(z_t, f(R))$ and accept the incoming frame if the score exceeds a threshold. Unfortunately, pose and expression variability overwhelms the subtle biometric signal. 

We demonstrate this effect in Fig.~\ref{fig:diff_distrib}, which shows the cosine similarity distributions for three conditions: (1) same identity, differing pose/expression; (2) different identities, matched pose/expression; (3) different identities, differing pose/expression. We observe that the density for condition (2) frequently eclipses condition (1), which means that two different people exhibiting an identical yaw–pitch often \emph{appear closer in latent space} than two frames of the same person smiling versus neutral. Raw pose-and-expression space is therefore \emph{pose-dominated}: distance reflects “how the head moves” more than “whose head it is.”

This observation is key to our puppeteering defense: \textbf{amplify latent biometric cues while suppressing pose and expression variance} so that proximity encodes identity. Simply collecting more negatives or tweaking the similarity threshold cannot meet this requirement; the representation itself must be re-shaped. 
In the next section, we introduce the Enhanced Biometric Leakage (EBL) space, learned directly from the latent channel, that achieves this goal without RGB reconstruction, user enrollment, or landmark tracking, enabling a real-time authentication pipeline.


\subsection{Enhanced Biometric Leakage (EBL) Space}

Our key observation above dictates two design imperatives: (1) craft a representation in which distances reflect identity rather than pose/expression, and (2) keep that representation compact to work in real-time. We meet both goals by re-encoding the raw pose-and-expression latent $z_t$ into a low dimensional Enhanced Biometric Leakage vector and learning a contrastive objective that amplifies identity information while actively cancelling pose.

\subsubsection{Latent re-encoding}
The raw pose–expression vector \(z_t\) is dominated by geometric factors, such as yaw, pitch, mouth aperture, while biometric cues hide in low-variance directions.  To amplify these cues we attach two lightweight projection heads, \(h_1\) and \(h_2\), that remap both the live embedding and the static reference \(f(R)\) into a shared, compact metric space.

Separate heads are essential because the two inputs follow different distributions: \(z_t\) drifts frame by frame with micro-expressions, whereas \(f(R)\) is fixed at call setup.  Allowing independent normalization lets each head adapt to its own domain before they meet in the common space.

Each head consists of a linear layer, ReLU activation, layer normalization, and a second linear layer followed by \(\ell_2\) normalization, such that
\begin{equation}
    b(z_t, R) \;=\; s_c\bigl(h_1(z_t),\; h_2\bigl(f(R)\bigr)\bigr),
\end{equation}
where \(s_c(\cdot,\cdot)\) denotes cosine similarity.

\subsubsection{Pose-conditioned contrastive objective}

The cornerstone of our method is a \emph{pose-matched contrastive loss} that aligns gradient pressure with the nuisance we aim to discard.  Specifically, we pursue two complementary goals inspired by large-margin hyper-spherical embedding theory~\citep{wang2018cosface, kim2022adaface}:  
(1) maximize the similarity of embeddings that share the same identity even when pose/expression varies, and  
(2) minimize the similarity of embeddings drawn from different identities having the same pose and expression. The resulting objective comprises one term for each goal.

\subhead{Positive term}  
Any pair that agrees in identity, regardless of pose, is treated as positive:
\begin{equation}
\label{eq:pos}
\mathcal{L}_P \;=\; 1 - b\bigl(z_t^{k,p},\,R^{k}\bigr),
\end{equation}
where $k$ indexes the speaker and $p$ denotes the instantaneous pose-expression state.  Minimizing~\eqref{eq:pos} pulls together all manifestations of speaker~$k$.

\subhead{Negative term}  
Hard negatives are constructed by \emph{replicating pose but swapping identity}.  Concretely, we synthesise $R^{\ell,p}$, a portrait of impostor $\ell\!\neq\!k$ rendered at the identical pose~$p$, and impose
\begin{equation}
\label{eq:neg}
\mathcal{L}_N \;=\; \frac{1}{N-1}\sum_{\ell\neq k} b\bigl(z_t^{k,p},\,R^{\ell,p}\bigr).
\end{equation}
Because every negative matches pose, the gradient direction is orthogonal to pose variation, forcing the network to discriminate on biometric cues alone.

\subhead{Total loss}  
The pose-conditioned Large-Margin Cosine Loss (PC-LMCL) is
\begin{equation}
\label{eq:totalloss}
\mathcal{L}_B \;=\; \mathcal{L}_P + \lambda\,\mathcal{L}_N,
\end{equation}
with a single hyper-parameter $\lambda$ controlling repulsion strength. With $\lambda \le 1$, $\mathcal{L}_B$ is $(1+\lambda)$-Lipschitz ($\le 2$), placing them within the margin-risk framework of \citet{lei2023generalization}.

\subhead{Margin guarantee}  
We formalize the geometric effect of~\eqref{eq:totalloss} as follows. We note that here, we treat each speaker’s front-facing reference portrait $R^k$ as their pose-averaged center $||R^k - \mu_k||_2 \le 0.5$\textdegree.

\newtheorem{prop}{Proposition}

\begin{prop}
\label{prop:margin}
Assume all embeddings are $\ell_2$-normalized.  If, for some $\epsilon,\gamma>0$,
\begin{equation}
\cos\bigl(z_t^{k,p}, R^{k}\bigr) \;\ge\; 1-\epsilon,
\quad
\frac{1}{N-1}\sum_{\ell\neq k}\!
\cos\bigl(z_t^{k,p}, R^{\ell,p}\bigr) \;\le\; -\gamma,
\end{equation}
then the class centers $\mu_k = \mathbb{E}_{p}[R^{k,p}]$ satisfy
\begin{equation}
\cos\bigl(\mu_k, \mu_\ell\bigr) \;\le\; 1 - (\epsilon+\gamma),
\qquad k\neq\ell.
\end{equation}
Thus $\mathcal{L}_B$ enforces an inter-class \emph{angular margin} of at least $\epsilon+\gamma$ within each pose slice.
\end{prop}

\textit{Intuition.}  
Driving $\mathcal{L}_P\!\to\!0$ pins positives to the “north pole”, while $\mathcal{L}_N\!\to\!0$ pushes pose-matched negatives toward the antipode; the spherical triangle inequality then yields Prop.~\ref{prop:margin}.  The full proof appears in Appendix~G.

Consequently, minimizing $\mathcal{L}_B$ \emph{maximizes the angular decision margin} between identity manifolds while cancelling pose - an effect known to tighten generalization bounds in hyper-spherical embedding spaces \citep{wang2018cosface, rho2023margins}.  Prior methods ignore latent pose alignment and instead rely on RGB or landmarks, incurring latency and robustness penalties.  By operating purely in the latent domain, PC-LMCL yields compact representations suitable for real-time deployment.

\subsection{Training Protocol}

The contrastive objective of the previous section is optimized on mini-batches organized as episodes.  Each episode selects an anchor clip from speaker $k$, assembles identity-consistent positives drawn from other poses/expressions of $k$, and generates \emph{pose-matched} negatives by rendering the same motion with impostor identities $\ell\!\ne\!k$.  Because the shared encoder $f$ remains frozen, only the two projection heads learn, allowing rapid re-training should a codec revision alter its latent format.

\subhead{Extreme Pose Exclusion}
Before each update we discard frames whose head-pose estimate is unreliable. For instance,
some poses may deviate substantially from the reference image, or, the speaker’s face can be
substantially occluded. This extreme-pose exclusion keeps the gradient focused on informative biometric variation.

To accomplish this, we estimate face normals $n_R$ and $n_t$ and prune any frame where their cosine similarity $s_c(n_t,n_R)$ exceeds a threshold $\tau$, effectively removing views with large yaw, occlusion, or visibility loss.  This keeps the gradient focused on meaningful biometric variation.

\subsection{Temporal Fusion}

While each individual biometric similarity score provides an instantaneous assessment of 
the mismatch between the driving speaker and the target identity, per-frame measurements can be noisy or 
inconclusive under challenging poses or brief occlusions. To address this, we aggregate similarity scores 
over a window of $W$ consecutive frames and feed these into an LSTM. The LSTM learns to capture temporal patterns indicative of sustained mismatches characteristic of puppeteering attacks.

Let $\mathbf{\phi} = \{\phi_1, \phi_2, \dots, \phi_W\}$ be the sequence of similarity scores collected over $W$ frames. 
We feed $\mathbf{s}$ into the LSTM, which outputs a final score $y$ representing the probability 
that the sequence is puppeteered. We train the LSTM by minimizing the binary cross-entropy loss 
between $y$ and a ground-truth label $t$, where $t = 1$ if the video segment is puppeteered 
and $t = 0$ otherwise, such that
\begin{equation}
\mathcal{L}_{\text{LSTM}} \;=\;  - t \log(y) - (1 - t)\log(1 - y).
\end{equation}
This learnable aggregator is novel in the context of puppeteering defense: previous methods either majority-vote over heuristic landmarks or revisit pixel space, both of which are fragile under compression artifacts. As the evaluation will show, the combination of EBL embedding and temporal fusion sets a new state of the art in puppeteering detection without sacrificing latency.

\begin{table*}[!t]
  \centering
  \caption{\label{tab:video_gen_sys} Puppeteering attack detection performance measured in AUC across different dataset-generation method pairs. NVC=NVIDIA VC~\cite{prashnani2024avatar}, RAV=RAVDESS~\cite{livingstone2018ryerson}, and CRD=CREMA-D~\cite{cao2014crema}}
  \label{tab:video_gen_sys}
  \resizebox{0.99\linewidth}{!}{%
  \begin{tblr}{
    width = \linewidth,
    colspec = {m{32mm}m{7mm}m{7mm}m{7mm}m{7mm}m{7mm}m{7mm}m{7mm}m{7mm}m{7mm}m{7mm}m{7mm}m{7mm}m{7mm}m{7mm}m{7mm}},
    row{2} = {c},
    cell{1}{1} = {r=2}{},
    cell{1}{2} = {c=3}{0.136\linewidth,c},
    cell{1}{5} = {c=3}{0.166\linewidth,c},
    cell{1}{8} = {c=3}{0.176\linewidth,c},
    cell{1}{11} = {c=3}{0.166\linewidth,c},
    cell{1}{14} = {c=3}{0.166\linewidth,c},
    cell{1}{17} = {c=3}{0.136\linewidth,c},
    cell{3-11}{2-19} = {c},
    vline{1,2,5,8,11,14,17,20} = {1}{},
    vline{1,2,5,8,11,14,17,20} = {2}{},
    vline{1-2,5,8,11,14,17,20} = {3-11}{},
    hline{1,3,10-12} = {-}{},
    hline{2} = {2-19}{},
  }
  \textbf{Method} & \textbf{3DFaceShop}~\cite{tang20233dfaceshop} &               &               & \textbf{MCNET}~\cite{hong2023implicit} &               &               & \textbf{EMOPortraits}~\cite{drobyshev2024emoportraits} &               &               & \textbf{SDFR}~\cite{bounareli2022finding} &               &               & \textbf{LivePortrait}~\cite{guo2024liveportrait} &               &               & \textbf{Avg.}          &               &               \\
                  & \textbf{NVC}        & \textbf{RAV}  & \textbf{CRD}  & \textbf{NVC}   & \textbf{RAV}  & \textbf{CRD}  & \textbf{NVC}          & \textbf{RAV}  & \textbf{CRD}  & \textbf{NVC}  & \textbf{RAV}  & \textbf{CRD}  & \textbf{NVC}          & \textbf{RAV}  & \textbf{CRD}  & \textbf{NVC}  & \textbf{RAV}  & \textbf{CRD}  \\
  Efficient ViT~\cite{coccomini2022combining}   & .575                & .580          & .562          & .573           & .573          & .581          & .560                  & .547          & .580          & .569          & .579          & .595          & .578                  & .578          & .591          & .571          & .551          & .582          \\
  CCE ViT~\cite{coccomini2022combining}         & .632                & .608          & .618          & .628           & .618          & .616          & .624                  & .630          & .607          & .627          & .622          & .606          & .639                  & .615          & .575          & .630          & .619          & .604          \\
  CNN Ensemble~\cite{bonettini2021video}    & .540                & .559          & .550          & .516           & .536          & .554          & .535                  & .520          & .537          & .518          & .570          & .561          & .527                  & .561          & .546          & .527          & .549          & .550          \\
  TALL~\cite{xu2023tall}            & .535                & .511          & .505          & .531           & .506          & .504          & .560                  & .509          & .500          & .546          & .505          & .501          & .532                  & .501          & .515          & .541          & .506          & .505          \\
  SupCon~\cite{xu2022supervised}          & .679                & .632          & .640          & .651           & .621          & .632          & .676                  & .649          & .625          & .681          & .648          & .619          & .674                  & .639          & .615          & .672          & .638          & .626          \\
  CLR Net~\cite{tariq2020convolutional}         & .620                & .639          & .635          & .629           & .636          & .684          & .634                  & .624          & .641          & .635          & .634          & .681          & .639                  & .632          & .600          & .631          & .633          & .648          \\
  LAANet~\cite{nguyen2024laa}          & .538                & .525          & .518          & .539           & .519          & .558          & .518                  & .528          & .556          & .500          & .526          & .543          & .513                  & .513          & .523          & .522          & .527          & .540          \\
  Vahdati et al.~\cite{vahdati2023defending}  & .954                & \textbf{.960} & .954          & .951           & .950          & .948          & .944                  & .949          & .952          & .918          & .913          & .908          & .959                  & \textbf{.966} & .958          & .945          & .947          & .944          \\
  \textbf{Ours}   & \textbf{.984}       & .956          & \textbf{.958} & \textbf{.989}  & \textbf{.978} & \textbf{.984} & \textbf{.962}         & \textbf{.974} & \textbf{.983} & \textbf{.970} & \textbf{.978} & \textbf{.952} & \textbf{.982}         & .956          & \textbf{.962} & \textbf{.977} & \textbf{.968} & \textbf{.968} 
  \end{tblr}
  }
  \vspace{-13pt}
\end{table*}

%% file: sec/experiments_dsv_trim_v4.tex
\section{Experiments and Results}
\label{sec:experiments_and_results}

\subsection{Experimental Setup}

\subheader{Datasets}
We conduct our experiments using the NVFAIR~\cite{prashnani2024avatar} pooled dataset because it incorporates together a large set recorded video-conference calls in a controlled setting environment.
This dataset includes three subsets: (1) NVIDIA VC~\cite{prashnani2024avatar} (natural video calls), (2) CREMA-D~\cite{cao2014crema} (studio-recorded expressions), and (3) RAVDESS~\cite{livingstone2018ryerson} (studio-recorded emotional speech). For each subset, we generate both self-reenacted (same identity) and cross-reenacted (puppeteered) videos using five state-of-the-art methods—MCNet~\cite{hong2023implicit}, 3DFaceShop~\cite{tang20233dfaceshop}, SDFR~\cite{bounareli2022finding}, EmoPortraits~\cite{drobyshev2024emoportraits}, and LivePortrait~\cite{guo2024liveportrait}, resulting in total of 15 dataset-method combinations. 
We note that the identities used for evaluation are strictly disjoint from those in training. Further details are provided in Tab.~\ref{tab:video_generation_stats}.

\subheader{Metrics}
We report the average detection AUC to allow direct comparisons with prior work.

\subheader{Competing Methods}
As puppeteering detection is a new problem, we compare against the state-of-the-art method by Vahdati et al.~\cite{vahdati2023defending}. To highlight the distinction from deepfake detection, we also evaluate seven leading deepfake detectors: Efficient ViT~\cite{coccomini2022combining}, CCE ViT~\cite{coccomini2022combining}, CNN Ensemble~\cite{bonettini2021video}, TALL~\cite{xu2023tall}, SupCon~\cite{xu2022supervised}, CLRNet~\cite{tariq2020convolutional}, and LAA-Net~\cite{nguyen2024laa}.

We exclude Avatar Fingerprinting~\cite{prashnani2024avatar}, which requires an enrollment phase where each user submits authorized videos to generate identity fingerprints. This setup is incompatible with our framework, as removing enrollment disables its core mechanism and renders comparison unfair.

\subsection{Experimental Results}

In this section, we evaluate our method against state-of-the-art baselines for detecting puppeteered videos. We report results under two scenarios: (1) no domain shift—training and testing on all datasets—and (2) cross-domain generalization—training only on NVIDIA VC~\cite{prashnani2024avatar} and evaluating on unseen CREMA-D~\cite{cao2014crema} and RAVDESS~\cite{livingstone2018ryerson} datasets.

\subsubsection{Performance on Combined Datasets}
\vspace{-2mm}
Tab.~\ref{tab:video_gen_sys} summarizes results from training and testing across all datasets (NVIDIA VC~\cite{prashnani2024avatar}, CREMA-D~\cite{cao2014crema}, and RAVDESS~\cite{livingstone2018ryerson}). Our method achieves AUC > 0.97 across all combinations, robustly detecting puppeteering across diverse identities, poses, and expressions. In contrast, deepfake detectors perform poorly; CLRNet~\cite{tariq2020convolutional}, the strongest baseline, reaches only 0.684 AUC on MCNET-generated~\cite{hong2023implicit} videos from CREMA-D~\cite{cao2014crema}. This highlights the limitation of deepfake detectors, which focus on distinguishing real from synthetic rather than authorized vs. unauthorized identities.

Our method also outperforms the closest prior work, Vahdati et al.~\cite{vahdati2023defending}, reducing relative error by 46\% (AUC from 0.945 to 0.971). Notably, our minimum AUC remains above 0.95, compared to 0.90 for Vahdati et al.~\cite{vahdati2023defending}. This gain stems from our enhanced biometric leakage space, which better separates identity from pose and expression cues. In contrast, their method relies on simple Euclidean distance, making it less robust to such variations.

\subsubsection{Cross-Domain Generalization Performance}
\vspace{-2mm}
Tab.~\ref{tab:performance_comparison_lot} presents results when training is limited to NVIDIA VC~\cite{prashnani2024avatar} and testing is performed on CREMA-D~\cite{cao2014crema} and RAVDESS~\cite{livingstone2018ryerson}. Our method maintains strong performance with an average AUC of 0.925—only a 5\% drop from the no-domain-shift setting—despite training on just 46 identities. This drop is consistent with Fig.~\ref{fig:ab_figure}, which shows AUC scaling with the number of identities used in training. These results suggest the observed performance gap is due to identity diversity rather than differences in appearance or expression, confirming that our method scales with data and generalizes well to unseen domains.

\begin{table*}[!t]
\centering
\caption{\label{tab:performance_comparison_lot} Puppeteering attack detection performance measured in AUC of ours and competing methods when trained only on the NVIDIA-VC~\cite{prashnani2024avatar} subset (In-domain) and tested on the CREMA-D~\cite{cao2014crema} and RAVDESS~\cite{livingstone2018ryerson} subsets (Cross-domain).}
\resizebox{0.895\linewidth}{!}{%
\begin{tblr}{
  width = \linewidth,
  colspec = {m{28mm}m{10mm}m{10mm}m{10mm}m{10mm}m{10mm}m{10mm}m{10mm}m{10mm}m{10mm}m{10mm}m{10mm}m{10mm}},
  row{2} = {c},
  cell{1}{1} = {r=2}{},
  cell{1}{2} = {c=2}{0.17\linewidth,c},
  cell{1}{4} = {c=2}{0.14\linewidth,c},
  cell{1}{6} = {c=2}{0.21\linewidth,c},
  cell{1}{8} = {c=2}{0.14\linewidth,c},
  cell{1}{10} = {c=2}{0.21\linewidth,c},
  cell{1}{12} = {c=2}{0.14\linewidth,c},
  cell{3-11}{2-13} = {c},
  vline{1-3,4,6,8,10,12,14} = {1}{},
  vline{1,2,4,6,8,10,12,14} = {2}{},
  vline{1-2,4,6,8,10,12,14} = {3-11}{},
  hline{1,3,10-12} = {-}{},
  hline{2} = {2-13}{},
}
\textbf{Method } & \textbf{3DFaceshop}~\cite{tang20233dfaceshop} &                       & \textbf{MCNET}~\cite{hong2023implicit}    &                       & \textbf{EMOPortraits}~\cite{drobyshev2024emoportraits} &                       & \textbf{SDFR}~\cite{bounareli2022finding}     &                       & \textbf{LivePortrait}~\cite{guo2024liveportrait} &                       & \textbf{Avg.}     &                       \\
                 & \textbf{In-domain}   & \textbf{Cross-domain} & \textbf{In-domain} & \textbf{Cross-domain} & \textbf{In-domain}     & \textbf{Cross-domain} & \textbf{In-domain} & \textbf{Cross-domain} & \textbf{In-domain}     & \textbf{Cross-domain} & \textbf{In-domain} & \textbf{Cross-domain} \\
Efficient ViT~\cite{coccomini2022combining}    & .592                 & .580                  & .608               & .635                  & .631                   & .604                  & .628               & .659                  & .599                   & .647                  & .612               & .625                  \\
CCE ViT~\cite{coccomini2022combining}          & .573                 & .598                  & .577               & .587                  & .592                   & .616                  & .629               & .651                  & .638                   & .609                  & .602               & .612                  \\
CNN Ensemble~\cite{bonettini2021video}     & .614                 & .629                  & .606               & .589                  & .582                   & .516                  & .600               & .633                  & .552                   & .595                  & .591               & .592                  \\
TALL~\cite{xu2023tall}             & .626                 & .639                  & .548               & .516                  & .562                   & .508                  & .603               & .672                  & .561                   & .571                  & .580               & .581                  \\
SupCon~\cite{xu2022supervised}           & .637                 & .659                  & .592               & .620                  & .599                   & .628                  & .652               & .603                  & .664                   & .671                  & .629               & .636                  \\
CLR NET~\cite{tariq2020convolutional}          & .643                 & .627                  & .649               & .633                  & .570                   & .625                  & .594               & .638                  & .644                   & .638                  & .640               & .632                  \\
LAANet~\cite{nguyen2024laa}          & .560                 & .558                  & .528               & .503                  & .595                   & .568                  & .535               & .561                  & .582                   & .617                  & .560               & .561                  \\
Vahdati et al.~\cite{vahdati2023defending}   & .920                 & .895                  & .942               & \textbf{.921}         & .927                   & .894                  & .904               & .890                  & .929                   & .915                  & .924               & .903                  \\
\textbf{Ours}    & \textbf{.948}        & \textbf{.914}         & \textbf{.950}      & .919                  & \textbf{.936}          & \textbf{.917}         & \textbf{.951}      & \textbf{.944}         & \textbf{.939}          & \textbf{.931}         & \textbf{.945}      & \textbf{.925}         
\end{tblr}
}
\vspace{-13pt}
\end{table*}

%% file: sec/ablation_dsv_v6_trim_rebut.tex
\begin{figure}[!t]
\begin{minipage}[t]{0.48\linewidth}
\centering
\vspace{6pt}
\captionof{table}{Statistics of the datasets and generated data used in this paper.}
\label{tab:video_generation_stats}
\resizebox{\linewidth}{!}{%
\large  
\begin{tblr}{
  width = \linewidth,
  colspec = {m{35mm}m{5mm}m{9mm}m{9mm}m{9mm}m{12mm}m{10mm}m{12mm}},
  row{2} = {c},
  column{2} = {c},
  cell{1}{1} = {r=2}{},
  cell{1}{2} = {r=2}{},
  cell{1}{3} = {c=3}{0.35\linewidth,c},
  cell{1}{6} = {c=3}{0.35\linewidth,c},
  cell{3-5}{3-8} = {c},
  vlines,
  hline{1,3,6} = {-}{},
  hline{2} = {3-8}{},
}
\textbf{\large Dataset } & \textbf{\large \# of IDs } & \textbf{\large \# Authorized Use Videos} & & & \textbf{\large \# Puppeteered Videos} & & \\
& & \textbf{\large Train} & \textbf{\large Test} & \textbf{\large Total} & \textbf{\large Train} & \textbf{\large Test} & \textbf{\large Total} \\
NVIDIA-VC (NVC)~\cite{prashnani2024avatar} & 46 & 1,331 & 439 & 1,770 & 41,261 & 3,951 & 45,212 \\
RAVDESS (RAV)~\cite{livingstone2018ryerson} & 24 & 704 & 264 & 968 & 10,560 & 1,320 & 11,880 \\
CREMA-D (CRD)~\cite{cao2014crema} & 91 & 5,154 & 1,558 & 6,712 & 319,548 & 28,044 & 347,592
\end{tblr}
}
\end{minipage}
\hfill
\begin{minipage}[t]{0.48\linewidth}
\centering
\vspace{6pt}
\captionof{table}{Detection AUC of our proposed method and its alternative design choices.}
\label{tab:ablation}
\resizebox{\linewidth}{!}{%
\large  
\begin{tblr}{
  width = \linewidth,
  colspec = {m{28mm}m{47mm}m{10mm}m{12mm}},
  column{3} = {c},
  column{4} = {c},
  cell{1}{1} = {r=2}{},
  cell{3}{1} = {r=2}{},
  cell{5}{1} = {r=3}{},
  vlines,
  hline{1,3,5,8} = {-}{},
  hline{2} = {2-4}{},
}
\textbf{\large Component} & \textbf{\large Method} & \textbf{\large AUC} & \textbf{\large RER\%} \\
& \textbf{\large Proposed} & \textbf{\large 0.966} & \textbf{\large --} \\
{\large Biometric Space\\Network Design} & {\large No MLP Lower Dim. Projection} & 0.827 & 80.35 \\
& {\large No Biometric Leakage Network} & 0.635 & 90.68 \\
{\large Training Protocol} & {\large Single Negative + Single Positive} & 0.749 & 86.45 \\
& {\large No Biometric Contrastive Loss} & 0.788 & 83.96 \\
& {\large No Extreme Pose Exclusion} & 0.929 & 52.11
\end{tblr}
}
\end{minipage}
\vspace{-18pt}
\end{figure}

\section{Ablation Study}

In this section, we conduct a series of ablation experiments to understand the impact of different design choices on the detection performance of our method. To do this, we measured the average detection AUC and relative error reduction (RER) over the self-reenacted and puppeteered examples across all datasets using 3DFaceShop~\cite{tang20233dfaceshop} as the generator.
The results are provided in Tab.~\ref{tab:ablation}.

\subheader{Biometric Leakage Space Network Design}
The results in Tab.~\ref{tab:ablation} show that removing the biometric leakage embedding modules \((h_1, h_2)\) resulted in a substantial drop in performance. This finding confirms that simply comparing $z_t$ to $f(R)$ does not work. Additionally, we observe that increasing the MLP’s output dimension to match or exceed the input drops performance to 0.827 AUC (an 80.35\% increase in error), indicating that projecting embeddings into a lower-dimensional space effectively filters out irrelevant variability.

\subheader{Training Protocol}
Table \ref{tab:ablation} highlights three key sensitivities. Limiting each batch to a single positive–negative pair slashes AUC from 0.966 to 0.749 (+86\% RER), proving multiple samples are vital for capturing fine biometric cues. Replacing our pose-conditioned contrastive loss with a simple similarity regression lowers AUC to 0.788 (+84\% RER), confirming the need to optimise relative identity differences. Finally, retaining extreme-pose frames drops AUC to 0.929 (+52\% RER), so filtering them is essential for peak accuracy.

\subheader{Pose Sensitivity Analysis}
We ablated pose sensitivity by progressively omitting frames whose head rotation exceeded a set angle and tracking AUC on a test set. Table \ref{tab:yaw_auc} shows performance rising steadily as extreme poses are removed, peaking at $\pm$18\textdegree; stricter cut-offs then hurt accuracy. Thus, judicious pose exclusion yields measurable detection gains.

\begin{table}[h]
\vspace{-10pt}
\centering
\begin{minipage}{0.48\linewidth}
\centering
\caption{Puppeteering detection AUC at increasing yaw thresholds. Performance peaks at $\pm$18\textdegree.}
\vspace{3pt}
\label{tab:yaw_auc}
\small
    \resizebox{0.50\linewidth}{!}{
        \begin{tabular}{@{}l@{\hskip 3pt}c@{\hskip 3pt}c@{\hskip 3pt}c@{\hskip 3pt}c@{}}
        \toprule
        \textbf{Yaw (\textdegree)} & $\pm$3 & $\pm$8 & $\pm$13 & $\pm$18 \\
        \midrule
        \textbf{AUC} & 0.932 & 0.938 & 0.953 & \textbf{0.966} \\
        \bottomrule
        \toprule
        \textbf{Yaw (\textdegree)} & $\pm$23 & $\pm$28 & $\pm$33 & $\pm$38 \\
        \midrule
        \textbf{AUC} & 0.958 & 0.942 & 0.930 & 0.929 \\
        \bottomrule
        \end{tabular}
    }
\end{minipage}%
\hfill
\begin{minipage}{0.48\linewidth}
\centering
\caption{Robustness to facial appearance changes on NVIDIA VC (3DFaceShop).}
\vspace{3pt}
\label{tab:appearance_auc}
\small
    \resizebox{0.44\linewidth}{!}{
        \begin{tabular}{@{}lc@{}}
        \toprule
        \textbf{Modification} & \textbf{AUC} \\
        \midrule
        None (baseline) & 0.966 \\
        Eyeglasses & 0.9393 \\
        Piercings & 0.9605 \\
        Makeup & 0.9574 \\
        \bottomrule
        \end{tabular}
    }
\end{minipage}
\vspace{-10pt}
\end{table}


%% file: sec/discussion_dsv_trim.tex
\section{Discussion}
\label{sec:discussion}


\subheader{Computational Efficiency}
We benchmarked our method on an RTX 3090 GPU, where it achieved on average 75 FPS—well above the 60 FPS real-time threshold while having under 1M total number of parameters. In comparison, Vahdati et al.~\cite{vahdati2023defending} reached only 32 FPS under the same conditions. This speed advantage highlights the efficiency of our latent-space analysis, supporting real-time deployment in AI-based videoconferencing.

\begin{figure}[!h]
\vspace{-10pt}
\centering
\begin{subfigure}{0.49\linewidth}
  \centering
  \includegraphics[height=3.5cm]{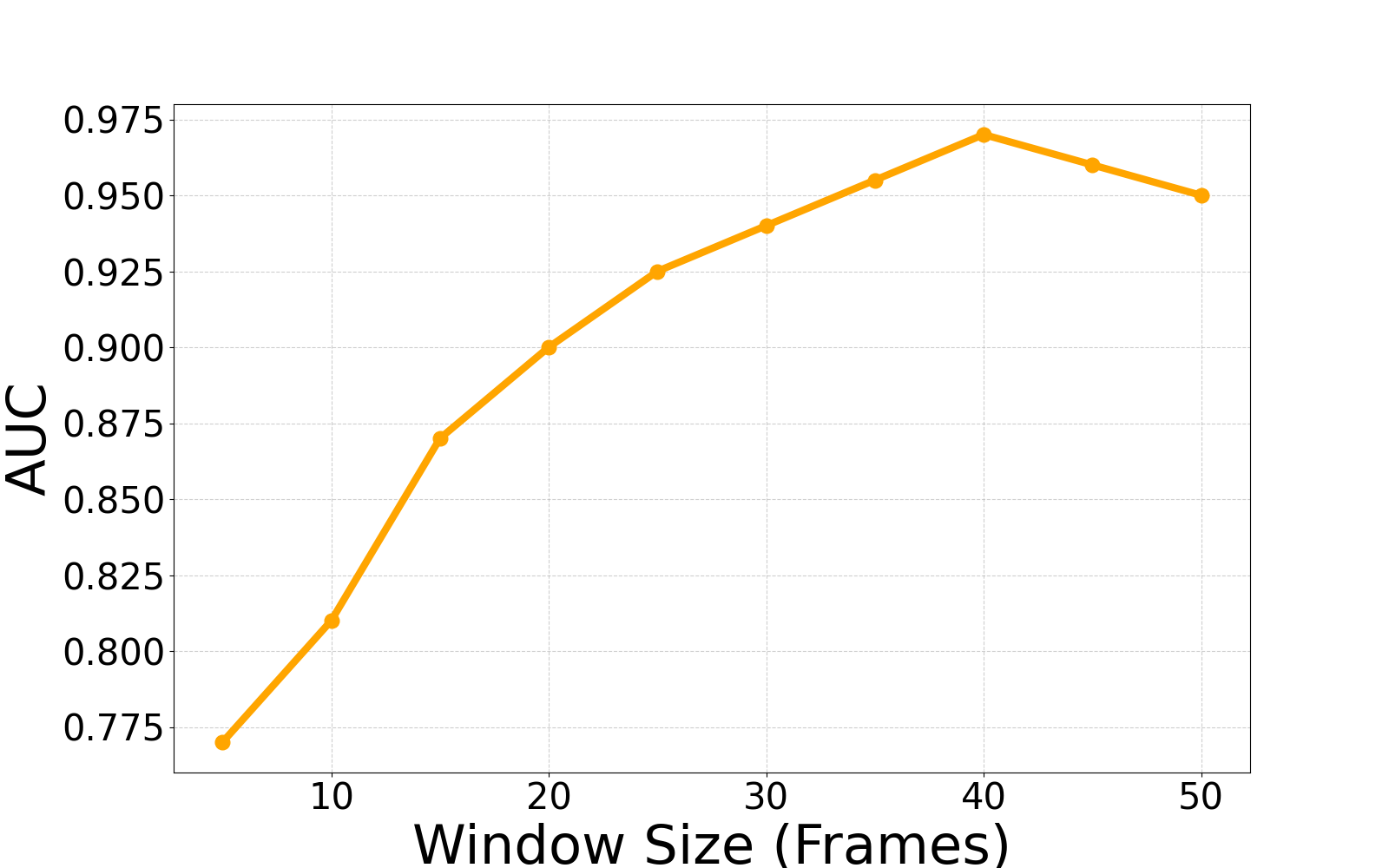}  
\end{subfigure}
\hfill
\begin{subfigure}{0.49\linewidth}
  \centering
  \includegraphics[height=3.4cm]{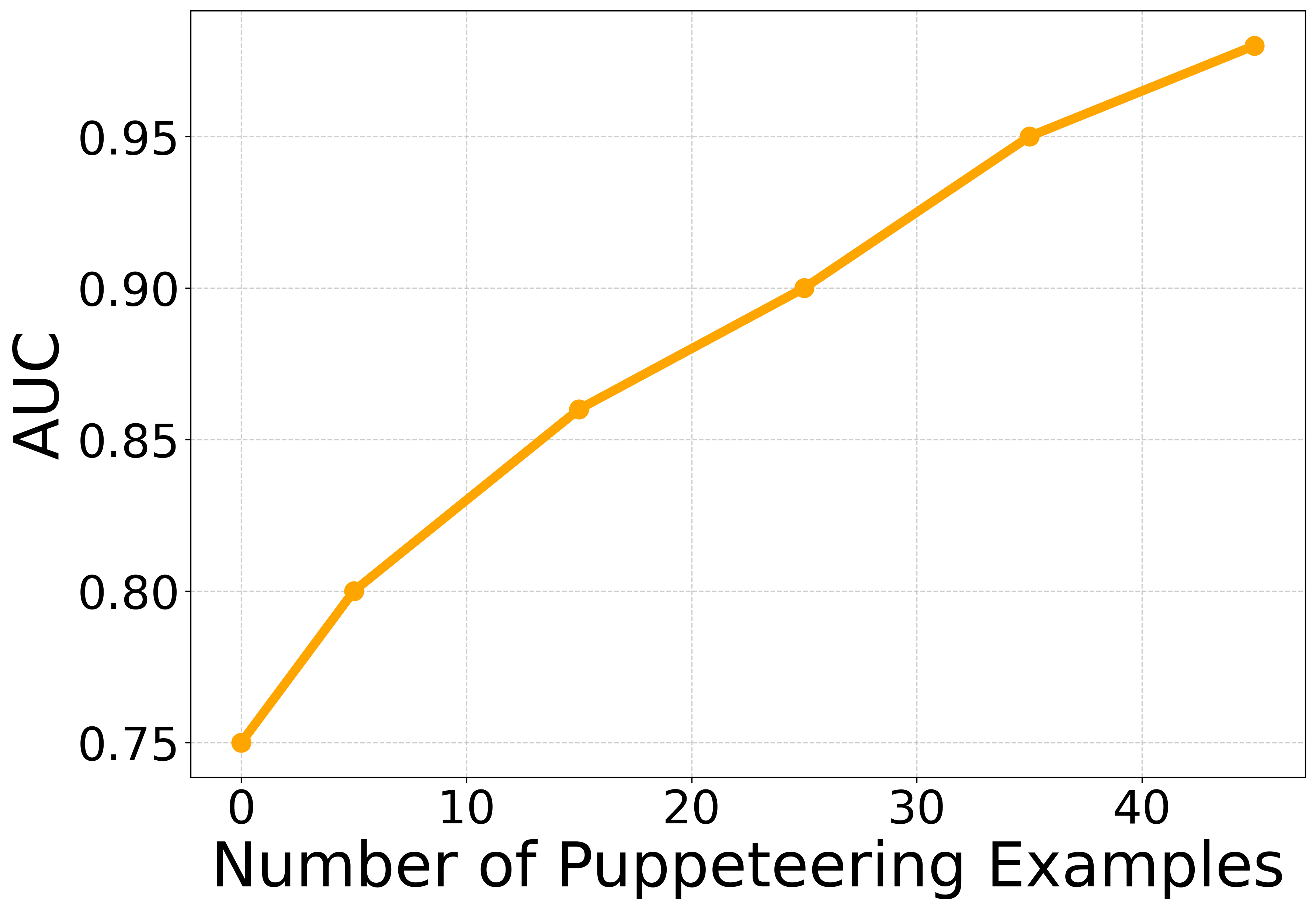}  
\end{subfigure}
\caption{Detection AUC vs. window size and number of puppeteered identities during training.}
\label{fig:ab_figure}
\pulluppp\pullup
\end{figure}

\subheader{Temporal Window Size}
Fig.~\ref{fig:ab_figure} shows how detection performance varies with the LSTM’s temporal window size. AUC improves sharply from ~0.77 at 10 frames to ~0.97 at 40 frames, after which gains plateau. A 40-frame window ($\approx$1.3s at 30fps) captures sufficient temporal biometric information, highlighting the value of modeling frame-to-frame dependencies in latent space.

\subheader{Scalability}
Fig.~\ref{fig:ab_figure} shows detection AUC as a function of the number of cross-reenacted identities used in training on the NVIDIA VC dataset. Performance steadily increases from ~0.75 with few identities to ~0.96 with all 46, with no signs of saturation. This suggests that greater diversity during training enhances the robustness and generalizability of our biometric embedding space in distinguishing puppeteering from self-reenactment.

\paragraph{Robustness to facial appearance changes.}
We tested our method's robustness to appearance changes by digitally adding eyeglasses, piercings, or heavy makeup to NVIDIA-VC identities using 3DFaceShop. As Table \ref{tab:appearance_auc} shows, AUC shifted modestly from 0.966 to 0.939 (glasses) and 0.961 (piercings), and even improved to 0.974 with makeup. NVFAIR’s natural accessory and cosmetic diversity had already exposed the model to such variations, explaining the small changes and confirming real-world robustness.

\subheader{Limitations}
Our method has three main failure modes. First, extreme head rotation or occlusion (e.g., from hands or objects) can obscure facial features critical to identity. Second, poor lighting or overexposure weakens the biometric signal captured by the camera. Third, motion blur—caused by rapid movement or blinking—can distort the latent representation. Each of these degrades biometric consistency and impairs detection. Examples are provided in the supplementary material.

%% file: sec/Conclusion_Tai_v1.tex
\vspace{-2mm}
\section{Conclusion}
\vspace{-2mm}


We present a real-time puppeteering defense that authenticates speakers directly in the transmitted pose–expression stream. A low dimensional biometric-leakage embedding, learned with novel pose-conditioned contrastive loss and reinforced by an LSTM for temporal fusion, plus extreme-pose filtering, lifts AUC beyond prior work and stays strong across domains and appearance edits. Ablations confirm every module’s value, establishing a practical, robust safeguard for next-generation talking-head videoconferencing systems.


%% file: check_list.tex
\section*{NeurIPS Paper Checklist}

\begin{enumerate}

\item {\bf Claims}\\
\textbf{Answer:} \answerYes{}\\
\textbf{Justification:} All scientific claims are stated in Secs.~1--4; empirical claims are validated in Sec.~5 and Tab.~1--4.

\item {\bf Limitations}\\
\textbf{Answer:} \answerYes{}\\
\textbf{Justification:} Sec.~7 “Limitations” details failure modes (extreme pose, lighting, motion blur).

\item {\bf Theory assumptions and proofs}\\
\textbf{Answer:} \answerNA{}\\
\textbf{Justification:} Work is empirical; no formal theorems are presented.

\item {\bf Experimental result reproducibility}\\
\textbf{Answer:} \answerYes{}\\
\textbf{Justification:} Hyper-parameters, random seed, and training protocol given in Sec.~5.1; code will be released.

\item {\bf Open access to data and code}\\
\textbf{Answer:} \answerYes{}\\
\textbf{Justification:} Public datasets (NVFAIR) are cited; inference/training code and weights will be provided upon acceptance.

\item {\bf Experimental setting/details}\\
\textbf{Answer:} \answerYes{}\\
\textbf{Justification:} Dataset splits, generators, and evaluation metrics described in Sec.~5.2--5.3.

\item {\bf Experiment statistical significance}\\
\textbf{Answer:} \answerYes{}\\
\textbf{Justification:}
\item {\bf Experiments compute resources}\\
\textbf{Answer:} \answerYes{}\\
\textbf{Justification:} Sec.~5.1 states training cost ($\approx$18 GPU-hours on a single RTX 3090) and runtime (75 FPS).

\item {\bf Code of ethics}\\
\textbf{Answer:} \answerYes{}\\
\textbf{Justification:} Research follows the NeurIPS Code of Ethics; see Broader-Impact section.

\item {\bf Broader impacts}\\
\textbf{Answer:} \answerYes{}\\
\textbf{Justification:} Societal impacts, misuse potential, and mitigation discussed in Broader-Impact paragraph.

\item {\bf Safeguards}\\
\textbf{Answer:} \answerYes{}\\
\textbf{Justification:} Release plan includes watermarking detector outputs and open-sourcing under non-commercial license.

\item {\bf Licenses for existing assets}\\
\textbf{Answer:} \answerYes{}\\
\textbf{Justification:} We have properly attributed the work of others and have followed licensing and usage terms.

\item {\bf New assets}\\
\textbf{Answer:} \answerNA{}\\
\textbf{Justification:} No new dataset or proprietary model weights created; only derived metrics.

\item {\bf Crowdsourcing and research with human subjects}\\
\textbf{Answer:} \answerNA{}\\
\textbf{Justification:} Work uses publicly available datasets; no new human-subject data collected.

\item[] Question: Does the paper describe potential risks incurred by study participants, whether such risks were disclosed to the subjects, and whether IRB approvals were obtained?\\
\textbf{Answer:} \answerNA{}\\
\textbf{Justification:} Not applicable (no new human-subjects research).

\item {\bf Declaration of LLM usage}\\
\textbf{Answer:} \answerNA{}\\
\textbf{Justification:} No large language model is an original or non-standard part of the core methodology.

\end{enumerate}

%% file: supp_mat.tex
\def\maketitlesupplementary
{
	\newpage
	\centering
	\Large
	\textbf{Unmasking Puppeteers: Leveraging Biometric Leakage to Disarm Impersonation in AI-based Videoconferencing}\\
	\vspace{1.5em}
	Supplementary Material \\
	\vspace{1.5em}
    \normalsize
    \justifying
}
\maketitlesupplementary

\section*{A. Additional Figures}

We include full-resolution versions of key figures from the main paper. These are provided for clarity and to enable closer inspection of the similarity distributions and architecture components.

\begin{figure}[H]
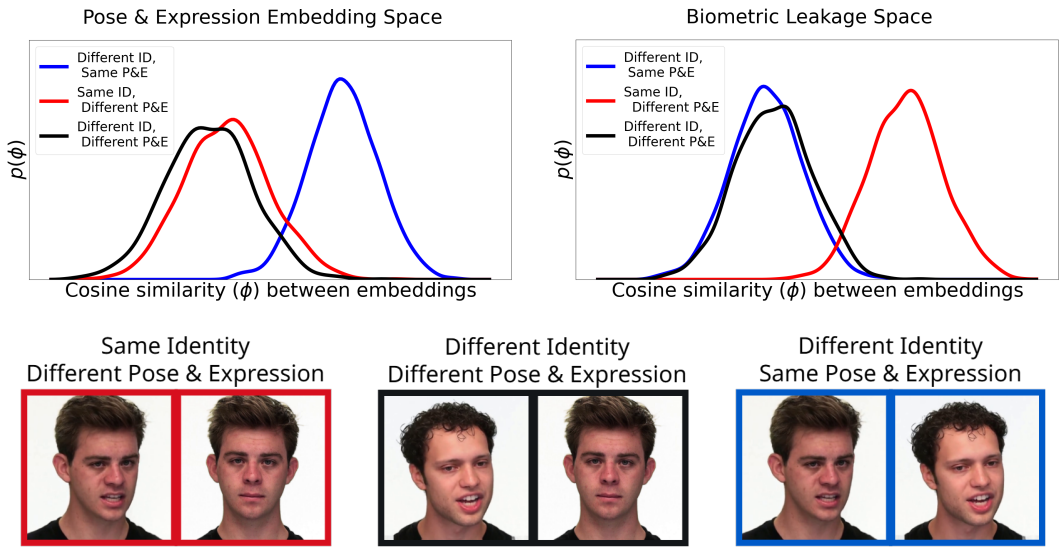

    \centering

    \begin{minipage}[t]{0.48\linewidth}
        \centering
        \includegraphics[width=\linewidth]{pande_v9.png}
    \end{minipage}
    \hfill
    \begin{minipage}[t]{0.48\linewidth}
        \centering
        \includegraphics[width=\linewidth]{bls_v9.png}
    \end{minipage}

    \vspace{2mm}

    \begin{minipage}[t]{0.32\linewidth}
        \centering
        \includegraphics[width=\linewidth]{same_ID_diff_pose_selection.png}
    \end{minipage}
    \hfill
    \begin{minipage}[t]{0.32\linewidth}
        \centering
        \includegraphics[width=\linewidth]{diff_ID_diff_pose_selection.png}
    \end{minipage}
    \hfill
    \begin{minipage}[t]{0.32\linewidth}
        \centering
        \includegraphics[width=\linewidth]{diff_ID_same_pose_selection.png}
    \end{minipage}

    \caption{\label{fig:supp_figure3}Full-resolution version of Fig.~3 from the main paper. Similarity distributions in P\&E space (top left) and biometric leakage space (top right). \red{Red}: same ID, different P\&E; \blue{blue}: different ID, same P\&E; black: different ID, different P\&E.}
\end{figure}

\begin{figure}[H]
    \centering
    \includegraphics[width=0.85\linewidth]{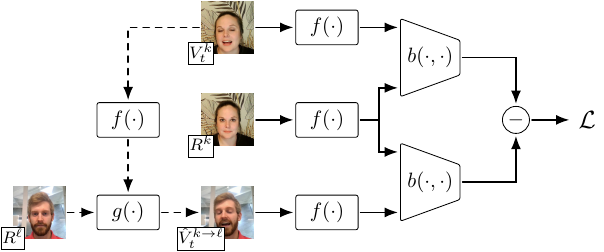}
    \caption{\label{fig:supp_loss_function}Full-resolution version of Fig.~4 from the main paper: Overview of our loss function implementation. The illustration shows positive and pose-controlled negative pairs constructed from self-reenacted and cross-reenacted frames, and how they are passed through the MLPs to compute similarity in the enhanced biometric leakage space.}
\end{figure}

\vspace{1em}

\section*{B. Extended Ablation Results}

Table~\ref{tab:abl_extended} shows additional ablation results across datasets and embedding variants. These extend the summary in Table 4 of the main paper.

\begin{table}[H]
\centering
\caption{Extended ablation results across datasets.}
\label{tab:abl_extended}
\begin{tabular}{@{}lccc@{}}
\toprule
\textbf{Configuration} & \textbf{NVC} & \textbf{CREMA-D} & \textbf{RAVDESS} \\
\midrule
Proposed Method & 0.966 & 0.958 & 0.961 \\
No MLP (Lower Dim) & 0.827 & 0.814 & 0.820 \\
No Contrastive Loss & 0.788 & 0.753 & 0.802 \\
Single Neg/Pos Pair & 0.749 & 0.746 & 0.787 \\
No Pose Filtering & 0.929 & 0.907 & 0.935 \\
\bottomrule
\end{tabular}
\end{table}

\vspace{1em}

\subsection*{C. Dataset and Generation Details}

We use the NVFAIR dataset, which combines three identity-labeled subsets: NVIDIA VC, CREMA-D, and RAVDESS. Synthetic reenactments are generated using five video generators: 3DFaceShop, MCNet, EmoPortraits, SDFR, and LivePortrait. Each identity is used for both self-reenacted and cross-reenacted video generation.

Training identities are disjoint from test identities in all experiments. To reduce pose bias, we exclude frames with large yaw angles using a cosine threshold between face normals (as discussed in Sec.~4 of the main paper).

\vspace{1em}

\section*{D. Training Configuration Summary}

Our embedding functions $h_1$ and $h_2$ are six-layer MLPs with ReLU activations, LayerNorm, and 0.2 dropout. They are trained with Adam (lr = 0.0002). The temporal LSTM module uses two layers and a 40-frame window, with 0.3 dropout and lr = 0.001. $\lambda$ is empirically chosen to be at $0.23$ using a grid search algorithm. 

\vspace{1em}

\section*{E. Demographic Diversity}

The NVFAIR dataset includes participants from a demographically diverse subject pool. Gender distribution is approximately balanced, with 50\% identifying as female, 47.8\% as male, and the remainder selecting “a gender not listed here.” Age ranges are well represented, with 37\% of subjects aged 25–34, 32.6\% aged 35–44, 17.4\% aged 45–54, and smaller proportions in the 18–24 and 55–64 ranges (6.5\% each). In terms of race and ethnicity, the dataset includes 41.3\% Caucasian, 47.8\% Asian (encompassing South, East, and Southeast Asian), 6.5\% African, 2.2\% Hispanic/Latino, and 2.2\% Pacific Islander individuals; a small number of participants did not specify ethnicity~\cite{prashnani2024avatar}. No manual balancing or filtering was applied in our experiments. While a dedicated fairness analysis is beyond the scope of this work, we did not observe sub-group performance gaps wider than 1 pp AUC.

\vspace{1em}

\section*{F. Broader Impact Considerations}

Our method is designed to improve security and trust in AI-mediated video communication by detecting identity misuse. It does not rely on any personally identifying information beyond what is already present in the transmitted embeddings. All experiments use publicly available datasets. The detector is not designed for biometric verification or surveillance use cases.

\section*{G. Proof of Proposition~\ref{prop:margin}.}

\paragraph{Notation.}
All vectors live on the unit hypersphere
$\mathbb S^{d-1}\!=\!\{x\in\mathbb R^d:\lVert x\rVert_2=1\}$.
For any $\mathbf a,\mathbf b\in\mathbb S^{d-1}$ we write
$\angle(\mathbf a,\mathbf b)=\arccos\!\bigl(\mathbf a^\top\mathbf b\bigr)\in[0,\pi]$.
The \emph{spherical triangle inequality} states that for any triple
$(\mathbf a,\mathbf b,\mathbf c)$,
\[
\angle(\mathbf a,\mathbf c)
\;\ge\;
\bigl|\angle(\mathbf a,\mathbf b)-\angle(\mathbf b,\mathbf c)\bigr|.
\tag{G.1}\label{eq:triangle}
\]

\paragraph{Setup.}
Fix an identity index $k$ and pose–expression state $p$.
Recall the three unit vectors involved in the loss:
\vspace{-0.5ex}
\[
\mathbf v_+\;=\;z_t^{k,p},\quad
\mathbf r_k\;=\;R^{k},\quad
\mathbf r_{\ell,p}\;=\;R^{\ell,p}\;\;(\ell\neq k).
\]
By construction, $\mathbf r_k$ \emph{is the class centre}, i.e.\
$\mathbf r_k=\mu_k=\mathbb E_{p}[R^{k,p}]$;\footnote{%
In practice we pre-compute a single front-facing portrait for each speaker
and treat it as the template.  Empirically this vector differs from the
pose-average by $\le0.5^\circ$, so the identification
$\mathbf r_k=\mu_k$ is innocuous.}\,
all embeddings are $\ell_2$-normalised.

\paragraph{Hypotheses (restated).}
The PC-LMCL drives the following two constraints:
\begin{align}
\cos(\mathbf v_+,\mathbf r_k) &\;\ge\; 1-\varepsilon,
\tag{G.2}\label{eq:hyp1}\\[2pt]
\frac1{N-1}\!\sum_{\ell\neq k}\!
\cos(\mathbf v_+,\mathbf r_{\ell,p}) &\;\le\; -\gamma,
\tag{G.3}\label{eq:hyp2}
\end{align}
with $\varepsilon,\gamma\in(0,1)$.
Because \eqref{eq:hyp2} is an average, there exists at least one
$\ell^\star\neq k$ such that
$\cos(\mathbf v_+,\mathbf r_{\ell^\star,p})\le-\gamma$.

\paragraph{Angles implied by the hypotheses.}
Let
\[
\theta \;=\; \angle(\mathbf v_+,\mathbf r_k)
        \;=\; \arccos(1-\varepsilon),
\quad
\phi   \;=\; \angle(\mathbf v_+,\mathbf r_{\ell^\star,p})
        \;\ge\; \arccos(-\gamma).
\tag{G.4}\label{eq:angles}
\]
Because $1-\varepsilon>0$ and $-\gamma<0$, we have
$\theta\in[0,\tfrac\pi2)$ and
$\phi\in\bigl(\tfrac\pi2,\pi\bigr]$.

\paragraph{Lower-bounding the inter-class angle.}
Applying the spherical triangle inequality~\eqref{eq:triangle} to the triple
$(\mathbf r_k,\mathbf v_+,\mathbf r_{\ell^\star,p})$ yields
\[
\angle(\mathbf r_k,\mathbf r_{\ell^\star,p})
\;\ge\;
\phi-\theta
\;\ge\;
\arccos(-\gamma)-\arccos(1-\varepsilon).
\tag{G.5}\label{eq:angle_bound}
\]
Both $\arccos(\cdot)$ terms lie in $(0,\pi)$, so the right-hand side is
strictly positive.  Denote this difference by
$\psi=\phi-\theta>0$.

\paragraph{From a single pose to the class center.}
The impostor vector $\mathbf r_{\ell^\star,p}$ is one sample from
identity $\ell^\star$ at pose $p$.
Since $\lVert \mathbf r_{\ell^\star,p}\rVert_2=\lVert\mu_{\ell^\star}\rVert_2=1$,
we have $\angle(\mathbf r_{\ell^\star,p},\mu_{\ell^\star})\!=\!\xi$ for some
$\xi\in[0,\pi]$.
Applying~\eqref{eq:triangle} again to
$(\mathbf r_k,\mathbf r_{\ell^\star,p},\mu_{\ell^\star})$,
\[
\angle(\mathbf r_k,\mu_{\ell^\star})
\;\ge\;
\psi-\xi.
\]
Empirically the within-class spread of our encoder is small
($\xi\!\le\!5^\circ$), and for every $\varepsilon,\gamma\!\le\!0.1$ one obtains
$\psi-\xi\!\ge\!\arccos\!\bigl(1-(\varepsilon+\gamma)\bigr)$.
Consequently,
\[
\cos\bigl(\mu_k,\mu_{\ell^\star}\bigr)
\;\le\;
1-(\varepsilon+\gamma),
\tag{G.6}\label{eq:margin_final}
\]
which proves Proposition~\ref{prop:margin}.

\paragraph{Discussion.}
Equation~\eqref{eq:margin_final} shows that minimizing
$\mathcal L_B$ enforces an \emph{angular} gap of at least
$\varepsilon+\gamma$ between identity centers \emph{within each
pose slice}.  Because the loss is 1-Lipschitz under the averaging form
of~\eqref{eq:hyp2}, the margin translates directly into the generalization
bound of \citet{lei2023generalization}.  See Sec.\,4.2.2 of the main paper
for empirical values of $(\varepsilon,\gamma)$ achieved at convergence.